\documentclass{article}

\usepackage{amsmath} 
\usepackage[ruled,vlined]{algorithm2e}
\usepackage{graphicx}
\usepackage{multirow}
\usepackage{amssymb}
\usepackage{colortbl}
\usepackage{wrapfig}
\newcommand{\up}[1]{{\color{darkgreen}$\uparrow$\,#1}}
\newcommand{\dn}[1]{{\color{red}$\downarrow$\,#1}}
 \usepackage[preprint]{neurips_2026}

\usepackage[utf8]{inputenc} % allow utf-8 input
\usepackage[T1]{fontenc}    % use 8-bit T1 fonts
\usepackage{hyperref}       % hyperlinks
\usepackage{url}           % simple URL typesetting
\usepackage{booktabs}       % professional-quality tables
\usepackage{amsfonts}       % blackboard math symbols
\usepackage{nicefrac}       % compact symbols for 1/2, etc.
\usepackage{microtype}      % microtypography
\usepackage{xcolor}         % colors
\definecolor{darkgreen}{rgb}{0.0, 0.55, 0.0}
\definecolor{rowgray}{gray}{0.9}
\title{LoMo: Local Modality Substitution for Deeper Vision-Language Fusion}

\definecolor{iccvblue}{rgb}{0.21,0.49,0.74}
\hypersetup{
  colorlinks=true,
  urlcolor=iccvblue
}
\urldef{\projecturl}\url{https://maplebb.github.io/LoMo}

\author{%
  \textbf{Feng Han$^{1,2}$} \quad
  \textbf{Zhixiong Zhang$^{2,3}$} \quad
  \textbf{Zheming Liang$^{2,4}$} \quad
  \textbf{Yibin Wang$^{1,2}$} \quad
  \textbf{Jiaqi Wang$^{2,5,*}$}\\
  \\[-0.1em]
  $^{1}$Fudan University \quad
  $^{2}$Shanghai Innovation Institute \quad
  $^{3}$Shanghai Jiao Tong University \\[-0em]
  $^{4}$University of Science and Technology of China \quad
  $^{5}$JD.COM\\
  \\[-0.1em]
  \projecturl
}

\begin{document}

\maketitle
\begingroup
\renewcommand{\thefootnote}{\fnsymbol{footnote}}
\footnotetext[1]{Corresponding authors.}
\endgroup

\begin{abstract}
Vision-Language Models (VLMs) have achieved substantial progress across a wide range of understanding and reasoning tasks, driven by large-scale image-text training aimed at multimodal fusion. Ideally, replacing a textual question with its rendered-image counterpart should leave model performance essentially unaffected. In practice, however, such modality substitution induces dramatic performance degradation. We attribute this ``carrier sensitivity'' issue to an inherent bias in current training corpora. Across prevalent datasets such as image captioning, VQA, OCR, and web-sourced interleaved data, text and images are typically organized into distinct and asymmetric roles, with text serving as linguistic queries and images as visual references. Such data bias leads VLMs to exhibit distinct preferences for information acquisition across different modalities. Consequently, VLMs fail to align representations of  semantically equivalent content across textual and visual carriers, making model reasoning fragile under modality substitution. To address this, we propose \textbf{Lo}cal \textbf{Mo}dality Substitution (\textbf{LoMo}), a lightweight, architecture-agnostic data curation paradigm designed to provide supervision for cross-modal representational invariance between semantically equivalent text and image carriers. LoMo achieves this by reformulating single-modality prompts into seamlessly interleaved multimodal sequences. It dynamically selects target text spans and recasts them as rendered images, thereby preserving the same semantics across ``text, visual, text'' carriers. Extensive experiments across 13 diverse multimodal benchmarks demonstrate that LoMo significantly improves overall multimodal reasoning and yields deeper cross-modal fusion. Specifically, it delivers consistent gains across foundational models, improving over standard SFT by 2.67 points on LLaVA-OneVision-1.5-8B and 2.82 points on Qwen3.5-9B.

\end{abstract}

\section{Introduction}
\label{sec:intro}

\begin{figure}[t]
  \centering
  \includegraphics[width=\textwidth]{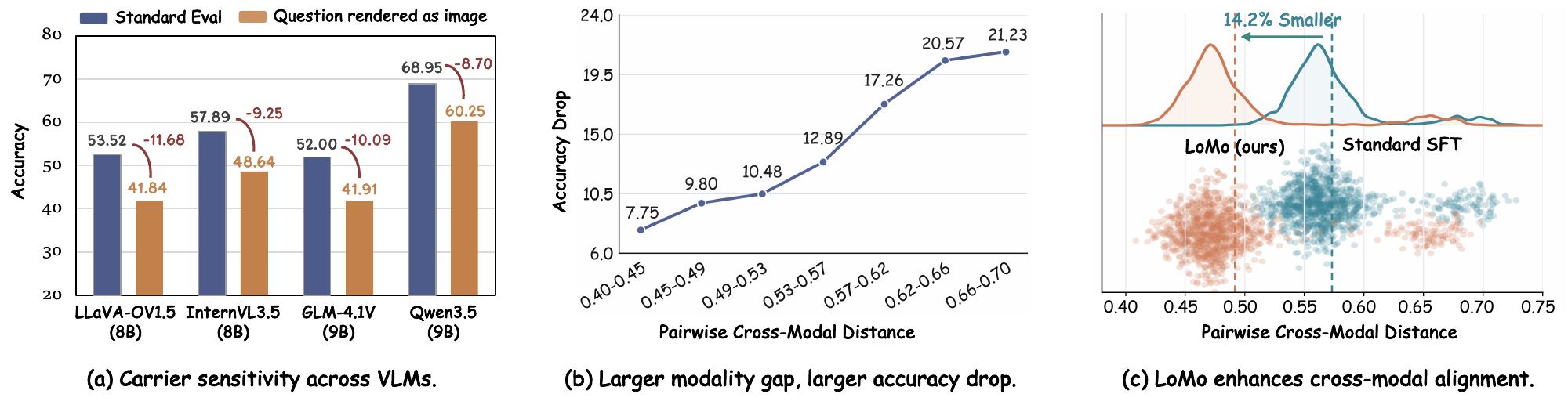}
  \caption{\textbf{Current Vision-Language Models exhibit carrier sensitivity driven by an underlying modality gap.} \textbf{(a) Carrier sensitivity across VLMs.} Simply shifting identical semantic content from a text format to a visual format (rendering standard questions as images) causes consistent and significant accuracy drops across state-of-the-art models. \textbf{(b) The physical manifestation of the modality gap.} By measuring the pairwise cross-modal distance between the original text and its rendered-image counterpart, we observe a strict monotonic trend, where greater representational distance between the two carriers corresponds to more severe accuracy degradation. \textbf{(c) LoMo enhances cross-modal alignment.} Our method shifts the cross-modal distance distribution markedly toward smaller values, reducing the average distance by \textbf{14.2\%} compared to Standard SFT and yielding tighter cross-carrier alignment.}
  \vspace{-12pt}
  \label{fig:teaser}
\end{figure}

Vision-Language Models (VLMs) have demonstrated strong generalization across diverse visual-language understanding tasks. Driven by rich image-text corpora and large-scale training aimed at multimodal fusion, state-of-the-art VLMs~\citep{an2025llava,bai2025qwen3,hong2025glm,wang2025internvl3,lu2024deepseek} exhibit powerful capabilities in tasks such as visual question answering, image captioning, document understanding, and visual grounding~\citep{liu2024mmbench,li2023seed,mathew2021docvqa}. Ideally, replacing the text of a multimodal query with its rendered-image counterpart should keep model performance largely stable. In practice, however, such modality substitution causes mainstream VLMs to suffer consistent and significant performance drops across multiple benchmarks, as shown in Figure~\ref{fig:teaser}(a). This exposes a severe \textbf{carrier sensitivity} problem. Although current VLMs process images and text jointly, their reasoning remains highly dependent on the modality carrier through which semantic content is presented. Merely switching identical semantics from a text carrier to a visual carrier can markedly degrade performance.

To trace this degradation to its source, we extract the hidden states of text inputs and their rendered-image counterparts, and measure their pairwise cosine distances. Grouping samples by this distance reveals a strict monotonic trend, where the average accuracy drop grows from 7.75\% in the closest bin to 21.23\% in the farthest (Figure~\ref{fig:teaser}(b)). This result indicates that the performance degradation is closely associated with a cross-carrier modality gap between semantically equivalent textual and visual inputs. We attribute this gap to an inherent bias in current multimodal training corpora. Across prevalent datasets such as image captioning, VQA, OCR, and web-sourced interleaved data, text and images are typically organized into distinct and asymmetric roles. Text often serves as linguistic instructions or queries, while images mainly provide visual references or evidence. Such data bias leads VLMs to exhibit distinct preferences for information acquisition across different modalities. Consequently, VLMs fail to align representations of semantically equivalent content across textual and visual carriers.

\begin{figure}[t]
  \centering
  \includegraphics[width=\textwidth]{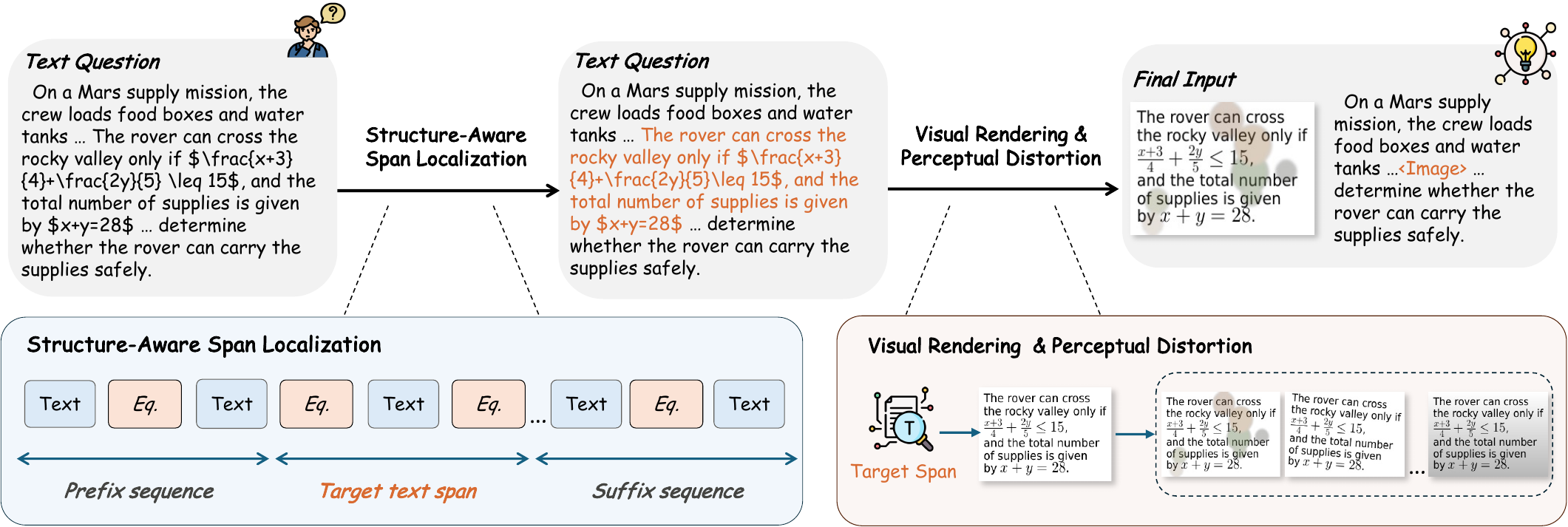}
  % \caption{te}
\caption{
\textbf{Overview of LoMo.} LoMo reformulates a text-only instance into a text–image interleaved sequence through three stages.
\textbf{Structure-Aware Span Localization} chunks the input in a formula-aware manner and selects a semantically coherent middle span as the target for visualization.
\textbf{Visual Rendering} converts the target span into an image via content-aware routing between LaTeX and standard text renderers. The image is then perturbed by \textbf{Perceptual Distortion} and substituted back into the original position, forming a ``text $\rightarrow$ visual carrier $\rightarrow$ text'' instance.
}
\label{fig:method}
\vspace{-12pt}
\end{figure}

Motivated by this, we propose \textbf{LoMo}, a lightweight and architecture-agnostic data curation paradigm designed to provide supervision for cross-modal representational invariance through local modality substitution, as shown in Figure~\ref{fig:method}. LoMo reformulates single-modality prompts into seamlessly interleaved multimodal sequences while preserving the original supervision target. In this way, the standard Supervised Fine-Tuning (SFT) objective is transformed into an implicit cross-carrier alignment signal that encourages the model to associate interleaved image-text inputs with their pure-text semantic counterparts. Specifically, LoMo consists of three sequential stages. \textbf{(1) Structure-Aware Span Localization} segments a text-only instance based on its semantic structure to identify target content for visualization. \textbf{(2) Visual Rendering}  recasts the selected span into a rendered visual carrier and embeds it between the surrounding text tokens, forming a ``text $\rightarrow$ visual $\rightarrow$ text'' sequence that promotes context-level fusion across modalities. \textbf{(3) Perceptual Distortion} applies real-world degradations to the visual carrier, ensuring that the learned fusion remains robust under perceptually challenging conditions. Crucially, LoMo is compatible with any multimodal training pipeline, requires no architectural modifications, introduces zero inference overhead, and demands no additional annotations.

\begin{figure}[t]
  \centering
  \includegraphics[width=\textwidth]{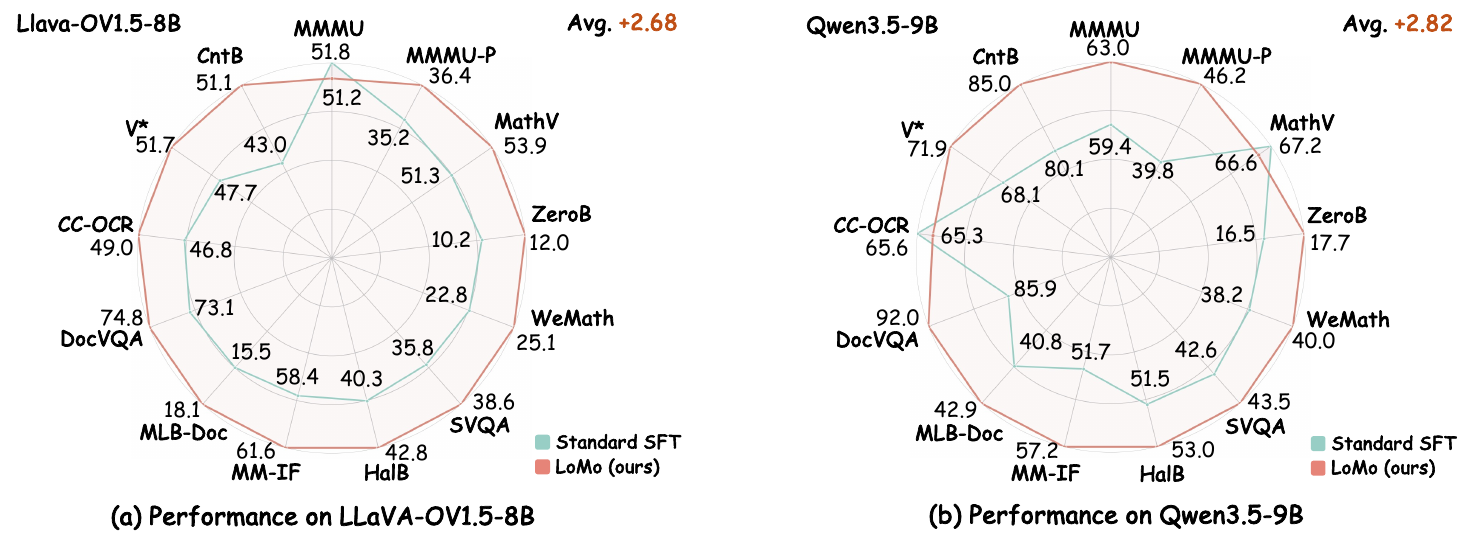}
  % \caption{te}
\caption{
LoMo yields consistent improvements over Standard SFT across two backbones (\textbf{(a)} +2.68 on LLaVA-OV1.5-8B; \textbf{(b)} +2.82 on Qwen3.5-9B over 13 benchmarks).
}
\vspace{-12pt}
\label{fig:radar}
  % \vspace{-8mm}
\end{figure}

Comprehensive experiments show that LoMo strengthens cross-modal fusion and delivers consistent gains across a wide spectrum of multimodal tasks. At the feature level, LoMo reduces the pairwise cross-modal distance by 14.2\% compared to the standard SFT model, indicating tighter cross-carrier alignment, as shown in Figure~\ref{fig:teaser}(c). Moreover, on 13 benchmarks spanning mathematical reasoning, VQA, OCR, document understanding, and visual perception, LoMo improves over the standard multimodal SFT baseline by 2.67 points on LLaVA-OneVision-1.5-8B and 2.82 points on Qwen3.5-9B, yielding stable improvements across backbones, as shown in Figure~\ref{fig:radar}. We further evaluate our method across data scales, where LoMo yields improvements in both downstream accuracy and representation-alignment metrics. Complementary analyses on the Modality Integration Rate~\citep{huang2024deciphering} further confirm that LoMo substantially enhances cross-modal fusion.

Our contributions are three-fold.
1) We systematically diagnose the \textbf{carrier sensitivity} problem in VLMs, revealing that it is closely associated with a cross-carrier modality gap induced by the distinct and asymmetric roles of text and images in standard training corpora.
2) We propose LoMo, a data-centric paradigm that performs local modality substitution to provide supervision for cross-modal representational invariance without architectural modifications or inference overhead.
3) We extensively validate LoMo on 13 multimodal benchmarks, demonstrating consistent accuracy improvements alongside improved cross-carrier representational consistency, with average gains of 2.67 and 2.82 on LLaVA-OneVision-1.5-8B and Qwen3.5-9B, respectively.

\section{Related Work}

\noindent \textbf{Vision-Language Models.}
Vision-language models (VLMs) extend LLMs to jointly process visual and
textual inputs, typically by aligning a pretrained vision encoder with
an LLM backbone. Architecturally, LLaVA~\citep{liu2023visual,liu2024improved}
established the simple ViT–MLP–LLM template, which has been scaled by
InternVL~\citep{chen2024internvl} and refined through systematic
exploration of vision encoders and connectors ~\citep{tong2024cambrian,tong2024eyes}. On the
training side, recent open-source families have improved data curation
and post-training: LLaVA-OneVision-1.5~\citep{an2025llava} restructures
the SFT corpus, Mantis~\citep{jiang2024mantis} reformats interleaved
multi-image instructions, and Insight-V~\citep{dong2025insight}
introduces long-chain visual reasoning data.
Qwen3-VL~\citep{bai2025qwen3},
InternVL3.5~\citep{wang2025internvl3}, and
GLM-4.1V-Thinking~\citep{hong2025glm} further push performance via
larger backbones and reinforcement learning. Despite their 
architectural and training-side advances, these recipes consistently 
treat text and images as modality-specific inputs, with text serving 
as instructions and images as visual scenes.

\noindent \textbf{Text-as-Pixels Modeling.}
In parallel, another line of 
work~\citep{xing2025vision,wang2024leveraging,kesen2025multilingual,cheng2025glyph,wei2025deepseek} has explored modeling text in pixel form rather
than as discrete tokens. Early efforts in OCR-free document
understanding, such as Pix2Struct~\citep{lee2023pix2struct}, learn to
parse rendered text through screenshot pretraining. Latent Compression Learning~\citep{yang2024vision} pushes this further by training vision encoders directly on web-scale image–text documents through a compression objective. More recently,
Glyph~\citep{cheng2025glyph} renders long documents into compact images
to extend the effective context window of VLMs, and
DeepSeek-OCR~\citep{wei2025deepseek} formalizes this idea as
\emph{contexts optical compression}, achieving high decoding accuracy at
$10\times$ token compression. A recent study~\citep{li2025text} further shows
that even off-the-shelf VLMs can read rendered text inputs with roughly
half the decoder tokens at little accuracy cost. These methods treat text-as-pixels as an efficiency-driven \emph{substitute} for text-as-tokens, aiming at OCR-style decoding or context compression. In contrast, our method treats text-as-pixels as a \emph{complement} to text-as-tokens within a single training instance, inducing an implicit cross-modal alignment supervision between the two carriers.

\noindent \textbf{Modality Gap and Cross-Modal Alignment.}
Aligning visual and textual representations remains a long-standing
challenge for multimodal models. The \emph{modality gap}~\citep{liang2022mind}
was first identified in CLIP-style models, where image and text embeddings
occupy disjoint regions of the shared space. Subsequent analysis~\citep{schrodi2024two}
traces this phenomenon to information imbalance between images and
captions, and shows that closing the gap can improve downstream
performance. Within decoder-based VLMs, the visual embedding space
inherited from CLIP has been shown to carry systematic blind spots
that propagate into MLLMs~\citep{tong2024eyes}, and the Modality
Integration Rate (MIR)~\citep{huang2024deciphering} reveals that a
measurable text–vision distribution gap persists in the shallow LLM
layers even after large-scale instruction tuning. The same
misalignment also drives multimodal hallucinations, motivating
decoding-time fixes such as VCD~\citep{leng2024mitigating} and
preference-optimization methods such as HA-DPO~\citep{sun2024aligning}.
These remedies operate at the decoding, or objective
level. In contrast, our method addresses the same gap from the data
side, reformulating text-only instances into
text$\rightarrow$visual$\rightarrow$text interleaved sequences so that
cross-carrier alignment becomes a task-level requirement during
standard SFT, with no architectural change and no inference overhead.

\section{Methodology}

\subsection{Overview and Formulation}
\label{sec:framework}

\textbf{Overview.} As discussed in Section~\ref{sec:intro}, current multimodal training paradigms lack explicit supervision for cross-modal representational invariance, leaving VLMs vulnerable to carrier sensitivity. To address this limitation, we propose \textbf{LoMo}, a data curation paradigm that provides an implicit cross-modal alignment supervision signal through local modality substitution. As illustrated in Figure~\ref{fig:method}, LoMo dynamically recasts a selected text span into a visual carrier through three successive stages. In \textbf{Structure-Aware Span Localization}, the input text is segmented into three parts, with the middle span identified as the target content. In \textbf{Visual Rendering}, the selected span is converted into images through a content-aware rendering pipeline. Finally, \textbf{Perceptual Distortion} applies semantics-preserving degradations to the rendered image, which is then substituted back into the position of the selected span, yielding a text–image interleaved instance. This reformulation is architecture-agnostic and compatible with any multimodal training pipeline, requiring no architectural changes, no additional annotations, and no inference overheads.

\textbf{Formulation.} 
Let $(x, a)$ denote an original text-only instance, where $x$ is the 
question and $a$ is the ground-truth answer. LoMo transforms $x$ 
through three successive operators, Structure-Aware Span Localization 
$\mathcal{S}(\cdot)$, Visual Rendering $\mathcal{R}(\cdot)$, and 
Perceptual Distortion $\mathcal{A}(\cdot)$. Formally,
\begin{equation}
(x_{\text{pre}},\, x_{\text{mid}},\, x_{\text{suf}}) = \mathcal{S}(x), 
\qquad I' = \mathcal{A}\big(\mathcal{R}(x_{\text{mid}})\big),
\end{equation}
which together produce the final mapping
\begin{equation}
\mathcal{T}(x) \triangleq (x_{\text{pre}},\, I',\, x_{\text{suf}}), \qquad (x, a) \;\longrightarrow\; \big((x_{\text{pre}},\, I',\, x_{\text{suf}}),\, a\big)
\end{equation}
The resulting instance forms a ``text $\rightarrow$ visual 
$\rightarrow$ text'' skeleton, requiring the model to jointly comprehend 
the surrounding textual context and the embedded visual carrier in 
order to recover the full semantics and predict $a$.

\subsection{Implementation of LoMo}
\label{sec:impl}

The carrier-substitution operator $\mathcal{T}(\cdot)$ is realized 
through three successive stages, jointly transforming a text-only 
instance $(x, a)$ into a text-image interleaved instance 
$(\mathcal{T}(x), a)$ while preserving the supervision target.

\textbf{Structure-Aware Span Localization} ($\mathcal{S}$) identifies 
a semantically coherent target span $x_{\text{mid}}$ for substitution. 
We first estimate the input length by sentence count. Short instances 
are taken entirely as $x_{\text{mid}}$ to fully preserve their semantic 
context, while long instances undergo a lightweight formula-aware chunking 
step that treats explicit mathematical expressions and common LaTeX 
commands as atomic, indivisible units. After chunking, the text $x$ 
is represented as an interleaved sequence of text and formula blocks,
\begin{equation}
x \mapsto ((t_1, l_1), (m_1, l_2), \dots, (t_n, l_{2n-1})),
\end{equation}
where $t$ and $m$ denote text and formula blocks respectively, and $l$ 
records the length of each block in characters. Guided by this 
representation, we extract the middle one-third of the sequence at 
block-level granularity as $x_{\text{mid}}$, ensuring that truncation 
boundaries never fall within an equation. The surrounding text 
$x_{\text{pre}}$ and $x_{\text{suf}}$ are retained, forming a 
``text $\rightarrow$ visual $\rightarrow$ text'' skeleton that 
compels the model to fuse both carriers in order to recover the full 
semantics and predict $a$.

\textbf{Visual Rendering} ($\mathcal{R}$) converts $x_{\text{mid}}$ 
into a rendered image through a content-aware routing pipeline that 
adapts to the properties of each span. Spans containing mathematical 
expressions are routed to a LaTeX-based renderer, which yields 
substantially more reliable formula typesetting than general-purpose 
text rendering, while spans without mathematical content are routed 
to a standard text-rendering pipeline. To safeguard throughput at 
scale, the renderer is wrapped in a fallback mechanism that 
automatically re-routes any LaTeX failure to the text renderer rather 
than discarding the instance. A mild margin-trimming step further 
removes large empty regions while preserving all rendered content, 
keeping image sizes bounded without altering their semantics.

\textbf{Perceptual Distortion} ($\mathcal{A}$) further perturbs each 
rendered image with semantics-preserving degradations, simulating the 
distortions document images commonly undergo during real-world capture 
and ensuring that the learned cross-carrier alignment is anchored to 
the underlying semantics. We define four sets of operations that jointly cover 
the range of perceptual noise observed in practical scenarios. 
\textit{Rotate} applies a large-angle or 
small-angle rotation to simulate orientation variations and slight 
tilt during capture. \textit{Blur} applies Gaussian, box, or motion 
blur to simulate camera shake. \textit{Shadow-or-stain} overlays edge 
shadows or surface stains to replicate uneven illumination and 
physical contamination, and \textit{Wave} induces local geometric 
deformations typical of folded paper or scanning artifacts. The final 
augmented image $I'$ is obtained by sampling one operation or by leaving the image 
unchanged.

\subsection{Implicit Cross-Modal Alignment Supervision of LoMo}
\label{sec:alignment}

We further examine how the local modality substitution of LoMo in 
Section \ref{sec:framework} reshapes the supervision signal of standard SFT. 
Standard SFT optimizes $f_\theta$ on each text-only instance $(x, a)$ 
through the negative log-likelihood
\begin{equation}
\mathcal{L}_{\text{SFT}}(\theta; x, a) = -\log p_\theta(a \mid x),
\label{eq:sft-loss}
\end{equation}
which constrains $f_\theta$ on the textual carrier $x$. LoMo augments 
this objective with an implicit cross-modal alignment signal through 
modality substitution, as we derive below.
\begin{equation}
\mathcal{L}_{\text{LoMo}}(\theta; x, a) 
= 
-\log p_\theta\!\big(a \,\big|\, \mathcal{T}(x)\big)
=
\underbrace{-\log p_\theta(a \mid x)}_{\text{standard SFT supervision}} 
\;+\; 
\underbrace{\log \frac{p_\theta(a \mid x)}{p_\theta\!\big(a \,\big|\, \mathcal{T}(x)\big)}}_{\text{cross-carrier alignment}}.
\label{eq:decomp}
\end{equation}
The first term recovers the standard SFT supervision in 
Eq.~\ref{eq:sft-loss}. To characterize the second term, we take the 
expectation of Eq.~\ref{eq:decomp} over $a \sim p_\theta(\cdot \mid x)$, 
under which the log-ratio reduces to a Kullback–Leibler divergence by 
definition, yielding
\begin{align}
\mathbb{E}_{a \sim p_\theta(\cdot \mid x)}\!\!\left[-\log p_\theta\!\big(a \,\big|\, \mathcal{T}(x)\big)\right] 
&= 
\mathbb{E}_{a \sim p_\theta(\cdot \mid x)}\!\!\left[-\log p_\theta(a \mid x)\right] 
+ 
\mathbb{E}_{a \sim p_\theta(\cdot \mid x)}\!\!\left[\log \frac{p_\theta(a \mid x)}{p_\theta\!\big(a \,\big|\, \mathcal{T}(x)\big)}\right] \notag \\
&= 
\mathbb{E}_{a \sim p_\theta(\cdot \mid x)}\!\!\left[-\log p_\theta(a \mid x)\right] 
+ 
\mathrm{KL}\!\Big(p_\theta(\cdot \mid x) \,\Big\Vert\, p_\theta\!\big(\cdot \,\big|\, \mathcal{T}(x)\big)\Big).
\label{eq:kl-decomp}
\end{align}
Optimizing $f_\theta$ on the carrier-substituted interleaved sequence 
is therefore equivalent to introducing an implicit cross-modal alignment term into the standard objective, driving the model's  predictive distributions on semantically equivalent textual and visual 
carriers toward agreement. This directly addresses the absence of cross-carrier representational invariance in current training paradigms.

\section{Experiments}
\label{sec:experiments}

\subsection{Experimental Setup}
\label{sec:setup}

\noindent \textbf{Models and training data.}
We examine LoMo on two open-source VLM backbones with substantially different architectures: \textbf{LLaVA-OneVision1.5-8B-Base}~\citep{an2025llava} and \textbf{Qwen3.5-9B-Base}~\citep{bai2025qwen3}. The training data is randomly sampled from the official LLaVA-OneVision1.5 SFT corpus~\citep{an2025llava}, comprising two million multimodal instruction examples and two million text-only instruction examples. The Standard SFT baseline directly fine-tunes on this pool. LoMo shares the same data pool, optimizer, learning-rate schedule, and number of training steps. The only difference is that a fraction of the text-only examples is reformatted into interleaved text-visual sequences. By construction, the two methods are matched in data scale, compute, and hyperparameters. Training hyperparameters and other implementation details of LoMo are provided in the Appendix~\ref{app:impl}.

\noindent \textbf{Evaluation benchmarks.}
We report results on 13 multimodal benchmarks spanning six categories. 
On general reasoning, we evaluate MMMU~\citep{yue2024mmmu} and MMMU-Pro~\citep{yue2025mmmu}. 
Math reasoning is covered by MathVista~\citep{lu2023mathvista}, ZeroBench~\citep{roberts2025zerobench}, and WeMath~\citep{qiao2025we}. We assess factuality with SimpleVQA~\citep{cheng2025simplevqa} and HallusionBench~\citep{guan2024hallusionbench}, and measure instruction following with MM-IFEval~\citep{ding2025mm}. Document and OCR understanding is probed via MMLongBench-Doc~\citep{ma2024mmlongbench}, DocVQA~\citep{mathew2021docvqa}, and CC-OCR~\citep{yang2025cc}. Finally, V$^*$~\citep{wu2024v} and CountBench~\citep{paiss2023teaching} target visual perception. All evaluations are conducted with EvalScope under identical prompting and decoding configurations.

\noindent \textbf{Evaluation protocols.}
We evaluate every benchmark under two protocols. \emph{Standard Evaluation} feeds the original (image, text question) pair to the model, matching standard practice. \emph{Rendered Evaluation} renders the entire text question as a single image, which replaces the original text and is fed to the model together with the original image. The linguistic content is identical across the two protocols and only the input modality differs. 

\noindent \textbf{Cross-modal alignment metrics.}
Beyond accuracy, we adopt two intrinsic metrics to probe the model's internal cross-modal alignment. \textit{(i) Modality Integration Rate (MIR)}~\citep{huang2024deciphering} quantifies the distributional gap between visual and textual tokens inside the VLM. Specifically, at each decoder layer the hidden states of visual and textual tokens are extracted and viewed as samples from two high-dimensional distributions, whose discrepancy is measured by the Fr\'echet Distance (FID). The per-layer FID computation follows the original paper. Since different backbones differ in the number of decoder layers, we report the layer-wise mean of FID as MIR. A lower MIR indicates a smaller distributional gap between textual and visual representations, reflecting tighter cross-modal integration. \textit{(ii) Pairwise Cross-Modal Distance} is a sample-level alignment metric. For each evaluation sample, we compute the mean hidden states of its text tokens and the corresponding rendered-image tokens at the output of the first VLM self-attention layer, denoted $\bar{h}_{\text{text}}$ and $\bar{h}_{\text{img}}$, and define their cosine distance as:
\begin{equation}
    d = 1 - \cos(\bar{h}_{\text{text}},\, \bar{h}_{\text{img}}).
\end{equation}
We average $d$ over the evaluation set; a lower value indicates that paired text and rendered image lie closer in representation space.

\begin{table*}[t]
\centering
\small
\setlength{\tabcolsep}{3.5pt}
\renewcommand{\arraystretch}{1.15}
\caption{Main results across 13 multimodal benchmarks under two evaluation protocols. \textbf{Standard Evaluation} feeds the original multimodal inputs (image + text question) to the model. \textbf{Rendered Evaluation} renders the entire text question as a single image. $\Delta$ denotes the absolute change of LoMo over Standard SFT; \textcolor{darkgreen}{$\uparrow$} and \textcolor{red}{$\downarrow$} indicate gains and drops. Benchmark abbreviations: MMMU-P: MMMU-Pro, MathV: MathVista, ZeroB: ZeroBench, SVQA: SimpleVQA, HalB: HallusionBench, MM-IF: MM-IFEval, MLB-Doc: MMLongBench-Doc, CntB: CountBench.}
\vspace{2mm}
\label{tab:main}
\resizebox{\textwidth}{!}{%
\begin{tabular}{l l cc ccc cc c ccc cc | c}
\toprule
\multirow{2}{*}{Model} & \multirow{2}{*}{Setting}
 & \multicolumn{2}{c}{Reasoning} & \multicolumn{3}{c}{Math}
 & \multicolumn{2}{c}{Factuality} & Instruct.
 & \multicolumn{3}{c}{Document \& OCR}
 & \multicolumn{2}{c|}{Visual Percept.} & \multirow{2}{*}{Avg.} \\
\cmidrule(lr){3-4}\cmidrule(lr){5-7}\cmidrule(lr){8-9}\cmidrule(lr){10-10}\cmidrule(lr){11-13}\cmidrule(lr){14-15}
 & & MMMU & MMMU-P & MathV & ZeroB & WeMath
 & SVQA & HalB & MM-IF & MLB-Doc & DocVQA & CC-OCR
 & V$^{*}$ & CntB & \\
\midrule
\rowcolor{rowgray}\multicolumn{16}{l}{\textit{Standard Evaluation}} \\
\multirow{3}{*}{LLaVA-OV1.5-8B}
 & Standard SFT
 & 51.78 & 35.24 & 51.30
 & 10.18 & 22.76
 & 35.51 & 40.35
 & 58.40 & 15.49 & 73.05
 & 46.76
 & 47.71 & 42.97
 & 40.88 \\
 & \textbf{+ LoMo}
 & \textbf{51.22} & \textbf{36.36} & \textbf{53.90}
 & \textbf{11.98} & \textbf{25.14}
 & \textbf{38.62} & \textbf{42.82}
 & \textbf{61.61} & \textbf{18.06} & \textbf{74.77}
 & \textbf{48.97}
 & \textbf{51.70} & \textbf{51.12}
 & \textbf{43.56} \\
 & $\Delta$
 & \dn{0.56} & \up{1.12} & \up{2.60}
 & \up{1.80} & \up{2.38}
 & \up{3.11} & \up{2.47}
 & \up{3.21} & \up{2.57} & \up{1.72}
 & \up{2.21}
 & \up{3.99} & \up{8.15}
 & \up{2.68} \\
\cmidrule(lr){1-16}
\multirow{3}{*}{Qwen3.5-9B}
 & Standard SFT
 & 59.44 & 39.83 & 67.20
 & 16.54 & 38.19
 & 42.57 & 51.53
 & 51.74 & 40.79 & 85.89
 & 65.61
 & 68.13 & 80.08
 & 54.43 \\
 & \textbf{+ LoMo}
 & \textbf{63.00} & \textbf{46.18} & \textbf{66.60}
 & \textbf{17.66} & \textbf{40.00}
 & \textbf{43.51} & \textbf{52.99}
 & \textbf{57.23} & \textbf{42.90} & \textbf{91.99}
 & \textbf{65.31}
 & \textbf{71.86} & \textbf{85.01}
 & \textbf{57.25} \\
 & $\Delta$
 & \up{3.56} & \up{6.35} & \dn{0.60}
 & \up{1.12} & \up{1.81}
 & \up{0.94} & \up{1.46}
 & \up{5.49} & \up{2.11} & \up{6.10}
 & \dn{0.30}
 & \up{3.73} & \up{4.93}
 & \up{2.82} \\
\midrule
\rowcolor{rowgray}\multicolumn{16}{l}{\textit{Rendered Evaluation}} \\
\multirow{3}{*}{LLaVA-OV1.5-8B}
 & Standard SFT
 & 21.22 & 16.01 & 18.10
 & 3.59 & 0.95
 & 22.42 & 41.36
 & 28.83 & 3.94 & 15.38
 & 14.49
 & 8.12 & 3.67
 & 15.24 \\
 & \textbf{+ LoMo}
 & \textbf{35.56} & \textbf{27.59} & \textbf{39.50}
 & \textbf{7.63} & \textbf{11.33}
 & \textbf{31.06} & \textbf{61.65}
 & \textbf{48.83} & \textbf{6.69} & \textbf{58.89}
 & \textbf{39.89}
 & \textbf{36.13} & \textbf{38.49}
 & \textbf{34.10} \\
 & $\Delta$
 & \up{14.34} & \up{11.58} & \up{21.40}
 & \up{4.04} & \up{10.38}
 & \up{8.64} & \up{20.29}
 & \up{20.00} & \up{2.75} & \up{43.51}
 & \up{25.40}
 & \up{28.01} & \up{34.82}
 & \up{18.86} \\
\cmidrule(lr){1-16}
\multirow{3}{*}{Qwen3.5-9B}
 & Standard SFT
 & 49.52 & 33.06 & 56.90
 & 15.94 & 23.43
 & 39.95 & 43.37
 & 41.25 & 28.87 & 47.24
 & 49.60
 & 61.58 & 71.66
 & 43.26 \\
 & \textbf{+ LoMo}
 & \textbf{62.48} & \textbf{45.29} & \textbf{65.50}
 & \textbf{16.62} & \textbf{39.72}
 & \textbf{43.26} & \textbf{47.05}
 & \textbf{54.32} & \textbf{36.57} & \textbf{91.73}
 & \textbf{66.14}
 & \textbf{64.73} & \textbf{83.98}
 & \textbf{55.18} \\
 & $\Delta$
 & \up{12.96} & \up{12.23} & \up{8.60}
 & \up{0.68} & \up{16.29}
 & \up{3.31} & \up{3.68}
 & \up{13.07} & \up{7.70} & \up{44.49}
 & \up{16.54}
 & \up{3.15} & \up{12.32}
 & \up{11.92} \\
\bottomrule
\end{tabular}}
\vspace{-12pt}
\end{table*}

\subsection{Main Results}
\label{sec:main_results}

Table~\ref{tab:main} reports performance under both evaluation protocols across all 13 benchmarks.

\noindent \textbf{Standard Evaluation.}
The upper block of Table~\ref{tab:main} reports Standard results on both backbones. LoMo consistently outperforms Standard SFT, with average gains of +2.68 on LLaVA-OneVision-8B and +2.82 on Qwen3.5-9B. The improvements are most pronounced on instruction following (MM-IFEval: +3.21\,/\,+5.49) and visual perception (CountBench: +8.15\,/\,+4.93; V$^*$: +3.99\,/\,+3.73), with consistent gains on factuality (SimpleVQA: +3.11\,/\,+0.94; HallusionBench: +2.47\,/\,+1.46), document \& OCR (DocVQA: +1.72\,/\,+6.10; MMLongBench-Doc: +2.57\,/\,+2.11), and math reasoning (WeMath: +2.38\,/\,+1.81). Overall, LoMo improves performance in 23 out of 26 comparisons, with only three marginal regressions, demonstrating its robust effectiveness across evaluation categories.

\noindent \textbf{Rendered Evaluation.}
The lower block of Table~\ref{tab:main} shows that the gap between LoMo and Standard SFT widens substantially when text is delivered through pixels. Average gains rise to \textbf{+18.86} on LLaVA-OneVision-8B and \textbf{+11.92} on Qwen3.5-9B, which is roughly $7\times$ and $4\times$ the corresponding Standard gains. The most dramatic improvements appear on document \& OCR (DocVQA: +43.51\,/\,+44.49; CC-OCR: +25.40\,/\,+16.54) and visual perception (CountBench: +34.82\,/\,+12.32; V$^*$: +28.01\,/\,+3.15). On Qwen3.5-9B, the Standard$\rightarrow$Rendered drop is compressed from \textbf{11.17} points under Standard SFT to just \textbf{2.07} under LoMo. In other words, models trained with Standard SFT collapse when the same linguistic content is delivered as pixels, whereas LoMo-trained models retain near-Standard performance.

We attribute the consistent gains under both protocols to LoMo's interleaved training format. By embedding rendered images between textual prefix and suffix, the model is repeatedly required to bridge text and pixels within a single sample. This fine-grained cross-modal fusion manifests as stronger visual perception and instruction following under Standard Evaluation, while also enabling more robust comprehension of pixel-rendered text under Rendered Evaluation. The key reason is that LoMo explicitly establishes correspondence between \emph{text-as-tokens} and \emph{text-as-pixels}, encouraging their representations to become more aligned. In contrast, Standard SFT largely preserves the functional separation between the two carriers, making the model sensitive to carrier substitution.

\begin{table*}[t]
\centering
\small
\setlength{\tabcolsep}{4pt}
\renewcommand{\arraystretch}{1.15}
\caption{Component ablation of LoMo on LLaVA-OV1.5-8B. \textit{Full-Text Rendering} renders the entire input as images without our Structure-Aware Span Localization or Perceptual Distortion. PD: Perceptual Distortion. Benchmark abbreviations follow Table~\ref{tab:main}. Best average is \textbf{bold}.}
\vspace{2mm}
\label{tab:ablation_components}
\resizebox{\textwidth}{!}{%
\begin{tabular}{l cc ccc cc c ccc cc | c}
\toprule
\multirow{2}{*}{Method}
 & \multicolumn{2}{c}{Reasoning} & \multicolumn{3}{c}{Math}
 & \multicolumn{2}{c}{Factuality} & Instruct.
 & \multicolumn{3}{c}{Document \& OCR}
 & \multicolumn{2}{c|}{Visual Percept.} & \multirow{2}{*}{Avg.} \\
\cmidrule(lr){2-3}\cmidrule(lr){4-6}\cmidrule(lr){7-8}\cmidrule(lr){9-9}\cmidrule(lr){10-12}\cmidrule(lr){13-14}
 & MMMU & MMMU-P & MathV & ZeroB & WeMath
 & SVQA & HalB & MM-IF & MLB-Doc & DocVQA & CC-OCR
 & V$^{*}$ & CntB & \\
\midrule
Standard SFT
 & 51.78 & 35.24 & 51.30
 & 10.18 & 22.76
 & 35.51 & 40.35
 & 58.40 & 15.49 & 73.05
 & 46.76
 & 47.71 & 42.97
 & 40.88 \\
+ Full-Text Rendering
 & 50.56 & 35.53 & 55.90
 & 11.68 & 22.76
 & 35.60 & 39.45
 & 59.51 & 16.87 & 71.80
 & 47.35
 & 47.32 & 52.55
 & 42.07 \\
+ LoMo w/o PD
 & 51.00 & 35.38 & 55.40
 & 11.98 & 25.33
 & 36.59 & 45.17
 & 60.00 & 16.59 & 73.77
 & 48.15
 & 50.65 & 50.31
 & 43.10 \\
\rowcolor{rowgray}+ LoMo
 & 51.22 & 36.36 & 53.90
 & 11.98 & 25.14
 & 38.62 & 42.82
 & 61.61 & 18.06 & 74.77
 & 48.97
 & 51.70 & 51.12
 & \textbf{43.56}\\
\bottomrule
\end{tabular}}
\vspace{-12pt}
\end{table*}

\noindent \textbf{Cross-modal Representation Analysis.}
We further evaluate cross-modal representation alignment using two complementary metrics: MIR captures \emph{set-level} alignment between the visual and textual token populations, while Paired Cross-Modal Distance captures \emph{pair-level} alignment. Both metrics start at the same values at initialization, but diverge sharply after training. At the 4M scale (Fig.~\ref{fig:data_scale}), LoMo reduces MIR by an additional 0.122 over Standard SFT, indicating a stronger global distributional alignment. The pair-level result is more striking: Standard SFT \emph{increases} the paired distance from 0.52 to 0.57, suggesting that conventional SFT pushes paired text and rendered-image representations apart at the sample level, whereas LoMo reduces it to 0.49.

This divergence highlights the distinction between set-level and pair-level alignment. In Standard SFT, text and images usually play complementary roles: text specifies the query, while images provide visual content. Thus, the model can optimize the training objective without explicitly aligning semantically equivalent text and rendered-image inputs. LoMo changes this training signal by splitting the required semantic cues between textual context and the rendered span. As a result, predicting the answer requires cross-carrier integration between \emph{text-as-tokens} and \emph{text-as-pixels}, turning alignment into a task-level requirement. This explains why LoMo improves both global distributional alignment and paired-sample alignment.

\subsection{Ablations}
\noindent \textbf{Component Ablation.}
Table~\ref{tab:ablation_components} compares three variants against Standard SFT: \textbf{Full-Text Rendering}, which renders the entire question as images without Structure-Aware Span Localization or Perceptual Distortion; \textbf{LoMo without Perceptual Distortion (LoMo w/o PD)}, which retains Structure-Aware Span Localization but skips Perceptual Distortion; and \textbf{LoMo}, which denotes the full pipeline. Naively rendering the full question yields only a +1.19 average gain, indicating that simple exposure to rendered text is insufficient when the rendered image is treated as an isolated visual input. In contrast, \textbf{LoMo w/o PD} already improves the average score by +2.22 over Standard SFT, showing that the Structure-Aware Span Localization is the dominant contributor. Meanwhile, adding Perceptual Distortion further raises the gain to +2.68, with the largest improvements appearing on visual perception and document understanding tasks, confirming the benefit of exposing the model to realistic visual carrier variations.

\begin{table*}[t]
\centering
\small
\setlength{\tabcolsep}{3.5pt}
\renewcommand{\arraystretch}{1.15}
\caption{\textbf{Quantitative results of different rewrite ratios on LLaVA-OV1.5-8B.} The rewrite ratio controls the fraction of text-only training samples reformatted into text--image interleaved sequences with LoMo. $\Delta$ denotes the average improvement over the Standard SFT baseline. Best average is \textbf{bold}.}
\vspace{2mm}
\label{tab:ratio_sweep}
\resizebox{\textwidth}{!}{%
\begin{tabular}{l c cc ccc cc c ccc cc | c c}
\toprule
\multirow{2}{*}{Setting} & \multirow{2}{*}{Rewrite Ratio}
 & \multicolumn{2}{c}{Reasoning} & \multicolumn{3}{c}{Math}
 & \multicolumn{2}{c}{Factuality} & Instruct.
 & \multicolumn{3}{c}{Document \& OCR}
 & \multicolumn{2}{c|}{Visual Percept.} & \multirow{2}{*}{Avg.} & \multirow{2}{*}{$\Delta$} \\
\cmidrule(lr){3-4}\cmidrule(lr){5-7}\cmidrule(lr){8-9}\cmidrule(lr){10-10}\cmidrule(lr){11-13}\cmidrule(lr){14-15}
 & & MMMU & MMMU-P & MathV & ZeroB & WeMath
 & SVQA & HalB & MM-IF & MLB-Doc & DocVQA & CC-OCR
 & V$^{*}$ & CntB & & \\
\midrule
Standard SFT & $0\%$
 & 51.78 & 35.24 & 51.30 & 10.18 & 22.76
 & 35.51 & 40.35 & 58.40 & 15.49 & 73.05 & 46.76
 & 47.71 & 42.97 & 40.88 & -- \\
\midrule
LoMo & $25\%$
 & 50.44 & 37.19 & 51.60 & 12.13 & 24.67
 & 37.33 & 42.84 & 62.58 & 17.14 & 74.52 & 49.84
 & 51.18 & 46.24 & 42.90 & \color{darkgreen}{+2.02} \\
\rowcolor{rowgray}LoMo & $50\%$ (ours)
 & 51.22 & 36.36 & 53.90
 & 11.98 & 25.14 & 38.62 & 42.82
 & 61.61 & 18.06 & 74.77 & 48.97
 & 51.70 & 51.12 & \textbf{43.56} & \color{darkgreen}{+2.68} \\
LoMo & $75\%$
 & 50.56 & 36.13 & 52.40 & 13.62 & 27.14
 & 37.23 & 44.25 & 61.22 & 17.51 & 74.74 & 49.19
 & 49.08 & 49.03 & 43.24 & \color{darkgreen}{+2.36} \\
LoMo & $100\%$
 & 50.67 & 36.32 & 53.80 & 11.23 & 26.38
 & 37.63 & 41.37 & 61.07 & 17.14 & 74.77 & 48.23
 & 50.07 & 46.18 & 42.68 & \color{darkgreen}{+1.80} \\
\bottomrule
\end{tabular}}
\vspace{-4pt}
\end{table*}

\begin{figure}[t]
  \centering
  \includegraphics[width=\textwidth]{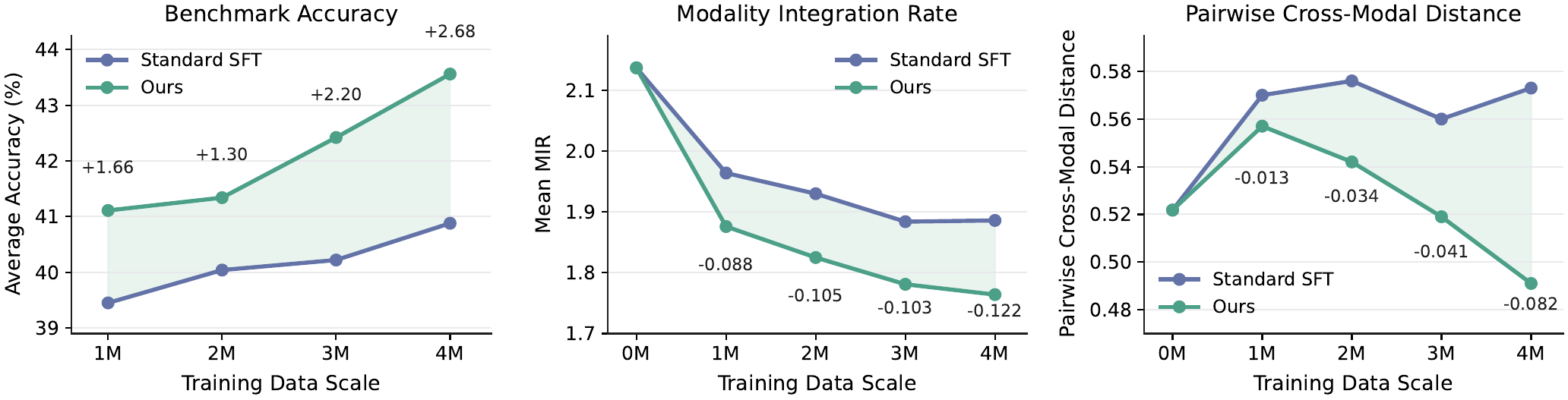}
  % \caption{te}
\caption{
\textbf{LoMo consistently outperforms standard SFT across data scales on three metrics.}
\textbf{(a)} Average accuracy on 13 multimodal benchmarks (higher is better).
\textbf{(b)} Mean MIR~\cite{huang2024deciphering}, measuring text-visual representation alignment (lower is better).
\textbf{(c)} Pairwise cross-modal distance, defined as $1 - \cos(\bar{h}_{\text{text}}, \bar{h}_{\text{img}})$ between hidden states of text and its rendered image (lower is better).
}
\label{fig:data_scale}
\vspace{-12pt}
  % \vspace{-8mm}
\end{figure}

% \noindent \textbf{Data Scale Ablation}
% To examine whether LoMo's advantage holds across data scales, we conduct experiments on four data scales. As shown in Figure~\ref{fig:data_scale}a, LoMo consistently outperforms Standard SFT at every scale, with gains increasing from +1.66 at 1M to +2.68 at 4M. The cross-modal alignment metrics, MIR and Paired Cross-Modal Distance, improve alongside accuracy, indicating that LoMo's benefits persist across different training data scales.
\noindent \textbf{Data Scale Ablation.}
To examine whether LoMo's advantage holds across data scales, we conduct experiments at four training scales (1M, 2M, 3M, 4M) and track three quantities along the same axis: average accuracy (Fig.~\ref{fig:data_scale}a), set-level alignment measured by MIR (Fig.~\ref{fig:data_scale}b), and pair-level alignment measured by Paired Cross-Modal Distance (Fig.~\ref{fig:data_scale}c). LoMo's accuracy gain over Standard SFT widens from +1.66 at 1M to +2.68 at 4M, and is accompanied by progressively stronger alignment on both metrics: at 4M, LoMo's MIR is 0.122 lower than Standard SFT, and its paired distance drops to 0.49, whereas Standard SFT's paired distance instead \emph{increases} from 0.52 to 0.57 over training. Across all four scales, LoMo improves both accuracy and cross-modal alignment over Standard SFT, indicating that its benefits are not specific to a particular training budget.

\noindent \textbf{Ablation on Rewrite Ratio.}
Quantitative comparisons are conducted on five rewrite ratios (0\%, 25\%, 50\%, 75\%, 100\%) on LLaVA-OV1.5-8B with other configurations fixed. Table~\ref{tab:ratio_sweep} reports results across the 13 multimodal benchmarks. All non-zero rewrite ratios yield substantial improvements over Standard SFT, confirming that LoMo consistently benefits multimodal fusion. Besides, the average accuracy first rises with the rewrite ratio, reaches $43.56$ at $50\%$, and then declines to $42.68$ at $100\%$. This trend suggests that the gain from cross-carrier supervision saturates at a moderate ratio, after which further rewriting yields diminishing returns.

\noindent \textbf{Ablation on Rendering Position.}
We compare four rendering positions on LLaVA-OV1.5-8B with all other configurations held fixed. \textbf{Prefix} and \textbf{Suffix} render the first or last one-third of the prompt, producing a single-image structure with text on one side. \textbf{Middle} renders the central one-third, producing a text--image--text structure that places the rendered span between two textual contexts. \textbf{Multi-Span} divides the prompt into multiple short segments and renders alternating segments, producing a text--image--text--image--text structure with multiple interleaving boundaries. Table~\ref{tab:position_sweep} reports results across the 13 multimodal benchmarks. Among these positions, Middle attains the highest average accuracy of $43.56$. The advantage of Middle over Prefix and Suffix indicates that placing the rendered image between two textual segments enforces stronger cross-carrier integration. Meanwhile, the result that Middle outperforms Multi-Span suggests that a single rendered span provides a more focused cross-carrier supervision signal than distributing renderings across multiple positions in the prompt.

\begin{table*}[t]
\centering
\small
\setlength{\tabcolsep}{3.5pt}
\renewcommand{\arraystretch}{1.15}
\caption{\textbf{Quantitative results of different rendering positions on LLaVA-OV1.5-8B.} The position controls where the rendered image is inserted into the original text sequence. \textit{Prefix} renders the first one-third of the prompt, yielding an image--text structure. \textit{Middle} (ours) renders the central one-third, yielding a text--image--text structure. \textit{Suffix} renders the last one-third. \textit{Multi-Span} renders two short spans, yielding a text--image--text--image--text structure. $\Delta$ denotes the average improvement over the Standard SFT baseline. Best average is \textbf{bold}.}
\vspace{2mm}
\label{tab:position_sweep}
\resizebox{\textwidth}{!}{%
\begin{tabular}{l c cc ccc cc c ccc cc | c c}
\toprule
\multirow{2}{*}{Setting} & \multirow{2}{*}{Position}
 & \multicolumn{2}{c}{Reasoning} & \multicolumn{3}{c}{Math}
 & \multicolumn{2}{c}{Factuality} & Instruct.
 & \multicolumn{3}{c}{Document \& OCR}
 & \multicolumn{2}{c|}{Visual Percept.} & \multirow{2}{*}{Avg.} & \multirow{2}{*}{$\Delta$} \\
\cmidrule(lr){3-4}\cmidrule(lr){5-7}\cmidrule(lr){8-9}\cmidrule(lr){10-10}\cmidrule(lr){11-13}\cmidrule(lr){14-15}
 & & MMMU & MMMU-P & MathV & ZeroB & WeMath
 & SVQA & HalB & MM-IF & MLB-Doc & DocVQA & CC-OCR
 & V$^{*}$ & CntB & & \\
\midrule
Standard SFT & --
 & 51.78 & 35.24 & 51.30 & 10.18 & 22.76
 & 35.51 & 40.35 & 58.40 & 15.49 & 73.05 & 46.76
 & 47.71 & 42.97 & 40.88 & -- \\
\midrule
LoMo & Prefix
 & 52.89 & 36.49 & 53.50 & 11.83 & 24.00
 & 35.85 & 40.75 & 59.06 & 15.77 & 74.60 & 47.98
 & 51.70 & 47.21 & 42.44 & \color{darkgreen}{+1.56} \\
\rowcolor{rowgray}LoMo & Middle (ours)
 & 51.22 & 36.36 & 53.90
 & 11.98 & 25.14 & 38.62 & 42.82
 & 61.61 & 18.06 & 74.77 & 48.97
 & 51.70 & 51.12 & \textbf{43.56} & \color{darkgreen}{+2.68} \\
LoMo & Suffix
 & 50.78 & 36.71 & 52.10 & 10.63 & 26.10
 & 37.58 & 40.20 & 60.64 & 15.12 & 75.58 & 49.59
 & 49.93 & 45.29 & 42.33 & \color{darkgreen}{+1.45} \\
LoMo & Multi-Span
 & 50.89 & 37.22 & 55.00 & 10.48 & 24.57
 & 37.48 & 41.49 & 62.01 & 16.32 & 75.30 & 49.72
 & 50.52 & 43.29 & 42.64 & \color{darkgreen}{+1.76} \\
\bottomrule
\end{tabular}}
\vspace{-12pt}
\end{table*}

\noindent \textbf{Controlled Comparison: Beyond More Image-Bearing Samples.}
\label{sec:matched_ratio}
LoMo converts a subset of originally text-only samples into image-bearing samples by rendering selected text spans, thereby increasing the effective number of image-bearing samples for training. To examine whether LoMo's gains simply come from this increased multimodal exposure, we resample the training data pool so that Standard SFT and LoMo share the same 1:1 ratio of image-bearing to text-only samples. Table~\ref{tab:matched_ratio} shows that LoMo still outperforms Standard SFT by 2.45 points on average under this matched setting. This indicates that LoMo's gains are driven by its interleaved cross-carrier formulation rather than by simply exposing the model to more image-bearing samples.

\begin{table*}[t]
\centering
\small
\setlength{\tabcolsep}{4pt}
\renewcommand{\arraystretch}{1.15}
\caption{Controlled comparison of LoMo under different image-bearing to text-only sample ratios on LLaVA-OV1.5-8B. The original setting uses LoMo's natural 3:1 ratio after rewriting, while the matched setting controls the effective ratio back to 1:1 to match Standard SFT for a fair comparison. $\Delta$ denotes the average improvement over Standard SFT. Best average is \textbf{bold}.}
\vspace{2mm}
\label{tab:matched_ratio}
\resizebox{\textwidth}{!}{%
\begin{tabular}{l c cc ccc cc c ccc cc | c c}
\toprule
\multirow{2}{*}{Setting} 
& \multirow{2}{*}{Image:Text Ratio}
& \multicolumn{2}{c}{Reasoning} 
& \multicolumn{3}{c}{Math}
& \multicolumn{2}{c}{Factuality} 
& Instruct.
& \multicolumn{3}{c}{Document \& OCR}
& \multicolumn{2}{c|}{Visual Percept.} 
& \multirow{2}{*}{Avg.}
& \multirow{2}{*}{$\Delta$} \\
\cmidrule(lr){3-4}\cmidrule(lr){5-7}\cmidrule(lr){8-9}\cmidrule(lr){10-10}\cmidrule(lr){11-13}\cmidrule(lr){14-15}
& 
& MMMU & MMMU-P 
& MathV & ZeroB & WeMath
& SVQA & HalB 
& MM-IF 
& MLB-Doc & DocVQA & CC-OCR
& V$^{*}$ & CntB 
& & \\
\midrule
Standard SFT
& 1:1
& 51.78 & 35.24
& 51.30 & 10.18 & 22.76
& 35.51 & 40.35
& 58.40
& 15.49 & 73.05 & 46.76
& 47.71 & 42.97
& 40.88 & -- \\
\midrule
\rowcolor{rowgray}LoMo
& 3:1 (Original)
& 51.22 & 36.36 
& 53.90 & 11.98 & 25.14
& 38.62 & 42.82
& 61.61
& 18.06 & 74.77 & 48.97
& 51.70 & 51.12
& \textbf{43.56} & \color{darkgreen}{+2.68} \\
LoMo
& 1:1 (Matched)
& 51.44 & 36.24
& 51.40 & 12.57 & 24.86
& 38.27 & 43.70
& 62.10
& 18.88 & 74.02 & 48.37
& 51.57 & 49.97
& 43.33 & \color{darkgreen}{+2.45} \\
\bottomrule
\end{tabular}}
\vspace{-12pt}
\end{table*}

\section{Conclusion}
In this work, we identify a carrier sensitivity phenomenon 
in current VLMs, where rendering identical textual content as a visual 
input causes consistent and substantial performance drops, with the 
magnitude tightly correlated to the cross-modal representational 
distance between the two carriers. We attribute this gap to the 
asymmetric roles of text and images in standard training corpora, and 
propose \textbf{LoMo}, a lightweight data curation paradigm that 
augments standard SFT with implicit cross-modal alignment 
supervision through local modality substitution. We hope LoMo offers 
a simple data-side recipe for bridging the modality gap and inspires 
further exploration of treating text and visuals as truly 
interchangeable semantic carriers.

\bibliographystyle{plain}
\bibliography{reference}

@article{huang2024deciphering,
  title={Deciphering cross-modal alignment in large vision-language models with modality integration rate},
  author={Huang, Qidong and Dong, Xiaoyi and Zhang, Pan and Zang, Yuhang and Cao, Yuhang and Wang, Jiaqi and Lin, Dahua and Zhang, Weiming and Yu, Nenghai},
  journal={arXiv preprint arXiv:2410.07167},
  year={2024}
}

@article{an2025llava,
  title={Llava-onevision-1.5: Fully open framework for democratized multimodal training},
  author={An, Xiang and Xie, Yin and Yang, Kaicheng and Zhang, Wenkang and Zhao, Xiuwei and Cheng, Zheng and Wang, Yirui and Xu, Songcen and Chen, Changrui and Zhu, Didi and others},
  journal={arXiv preprint arXiv:2509.23661},
  year={2025}
}

@article{bai2025qwen3,
  title={Qwen3-vl technical report},
  author={Bai, Shuai and Cai, Yuxuan and Chen, Ruizhe and Chen, Keqin and Chen, Xionghui and Cheng, Zesen and Deng, Lianghao and Ding, Wei and Gao, Chang and Ge, Chunjiang and others},
  journal={arXiv preprint arXiv:2511.21631},
  year={2025}
}

@inproceedings{yue2024mmmu,
  title={Mmmu: A massive multi-discipline multimodal understanding and reasoning benchmark for expert agi},
  author={Yue, Xiang and Ni, Yuansheng and Zhang, Kai and Zheng, Tianyu and Liu, Ruoqi and Zhang, Ge and Stevens, Samuel and Jiang, Dongfu and Ren, Weiming and Sun, Yuxuan and others},
  booktitle={Proceedings of the IEEE/CVF conference on computer vision and pattern recognition},
  pages={9556--9567},
  year={2024}
}

@inproceedings{yue2025mmmu,
  title={Mmmu-pro: A more robust multi-discipline multimodal understanding benchmark},
  author={Yue, Xiang and Zheng, Tianyu and Ni, Yuansheng and Wang, Yubo and Zhang, Kai and Tong, Shengbang and Sun, Yuxuan and Yu, Botao and Zhang, Ge and Sun, Huan and others},
  booktitle={Proceedings of the 63rd Annual Meeting of the Association for Computational Linguistics (Volume 1: Long Papers)},
  pages={15134--15186},
  year={2025}
}

@article{lu2023mathvista,
  title={Mathvista: Evaluating mathematical reasoning of foundation models in visual contexts},
  author={Lu, Pan and Bansal, Hritik and Xia, Tony and Liu, Jiacheng and Li, Chunyuan and Hajishirzi, Hannaneh and Cheng, Hao and Chang, Kai-Wei and Galley, Michel and Gao, Jianfeng},
  journal={arXiv preprint arXiv:2310.02255},
  year={2023}
}

@article{roberts2025zerobench,
  title={Zerobench: An impossible visual benchmark for contemporary large multimodal models},
  author={Roberts, Jonathan and Taesiri, Mohammad Reza and Sharma, Ansh and Gupta, Akash and Roberts, Samuel and Croitoru, Ioana and Bogolin, Simion-Vlad and Tang, Jialu and Langer, Florian and Raina, Vyas and others},
  journal={arXiv preprint arXiv:2502.09696},
  year={2025}
}

@inproceedings{qiao2025we,
  title={We-math: Does your large multimodal model achieve human-like mathematical reasoning?},
  author={Qiao, Runqi and Tan, Qiuna and Dong, Guanting and MinhuiWu, MinhuiWu and Sun, Chong and Song, Xiaoshuai and Wang, Jiapeng and Gongque, Zhuoma and Lei, Shanglin and Zhang, Yifan and others},
  booktitle={Proceedings of the 63rd Annual Meeting of the Association for Computational Linguistics (Volume 1: Long Papers)},
  pages={20023--20070},
  year={2025}
}

@inproceedings{cheng2025simplevqa,
  title={Simplevqa: Multimodal factuality evaluation for multimodal large language models},
  author={Cheng, Xianfu and Zhang, Wei and Zhang, Shiwei and Yang, Jian and Guan, Xiangyuan and Wu, Xianjie and Li, Xiang and Zhang, Ge and Liu, Jiaheng and Mai, Yuying and others},
  booktitle={Proceedings of the IEEE/CVF International Conference on Computer Vision},
  pages={4637--4646},
  year={2025}
}

@inproceedings{guan2024hallusionbench,
  title={Hallusionbench: an advanced diagnostic suite for entangled language hallucination and visual illusion in large vision-language models},
  author={Guan, Tianrui and Liu, Fuxiao and Wu, Xiyang and Xian, Ruiqi and Li, Zongxia and Liu, Xiaoyu and Wang, Xijun and Chen, Lichang and Huang, Furong and Yacoob, Yaser and others},
  booktitle={Proceedings of the IEEE/CVF conference on computer vision and pattern recognition},
  pages={14375--14385},
  year={2024}
}

@inproceedings{ding2025mm,
  title={Mm-ifengine: Towards multimodal instruction following},
  author={Ding, Shengyuan and Wu, Shenxi and Zhao, Xiangyu and Zang, Yuhang and Duan, Haodong and Dong, Xiaoyi and Zhang, Pan and Cao, Yuhang and Lin, Dahua and Wang, Jiaqi},
  booktitle={Proceedings of the IEEE/CVF International Conference on Computer Vision},
  pages={1099--1109},
  year={2025}
}

@article{ma2024mmlongbench,
  title={Mmlongbench-doc: Benchmarking long-context document understanding with visualizations},
  author={Ma, Yubo and Zang, Yuhang and Chen, Liangyu and Chen, Meiqi and Jiao, Yizhu and Li, Xinze and Lu, Xinyuan and Liu, Ziyu and Ma, Yan and Dong, Xiaoyi and others},
  journal={Advances in Neural Information Processing Systems},
  volume={37},
  pages={95963--96010},
  year={2024}
}

@inproceedings{mathew2021docvqa,
  title={Docvqa: A dataset for vqa on document images},
  author={Mathew, Minesh and Karatzas, Dimosthenis and Jawahar, CV},
  booktitle={Proceedings of the IEEE/CVF winter conference on applications of computer vision},
  pages={2200--2209},
  year={2021}
}

@inproceedings{yang2025cc,
  title={Cc-ocr: A comprehensive and challenging ocr benchmark for evaluating large multimodal models in literacy},
  author={Yang, Zhibo and Tang, Jun and Li, Zhaohai and Wang, Pengfei and Wan, Jianqiang and Zhong, Humen and Liu, Xuejing and Yang, Mingkun and Wang, Peng and Bai, Shuai and others},
  booktitle={Proceedings of the IEEE/CVF International Conference on Computer Vision},
  pages={21744--21754},
  year={2025}
}

@inproceedings{wu2024v,
  title={V?: Guided visual search as a core mechanism in multimodal llms},
  author={Wu, Penghao and Xie, Saining},
  booktitle={Proceedings of the IEEE/CVF Conference on Computer Vision and Pattern Recognition},
  pages={13084--13094},
  year={2024}
}

@inproceedings{paiss2023teaching,
  title={Teaching clip to count to ten},
  author={Paiss, Roni and Ephrat, Ariel and Tov, Omer and Zada, Shiran and Mosseri, Inbar and Irani, Michal and Dekel, Tali},
  booktitle={Proceedings of the IEEE/CVF international conference on computer vision},
  pages={3170--3180},
  year={2023}
}

@article{wang2025internvl3,
  title={Internvl3. 5: Advancing open-source multimodal models in versatility, reasoning, and efficiency},
  author={Wang, Weiyun and Gao, Zhangwei and Gu, Lixin and Pu, Hengjun and Cui, Long and Wei, Xingguang and Liu, Zhaoyang and Jing, Linglin and Ye, Shenglong and Shao, Jie and others},
  journal={arXiv preprint arXiv:2508.18265},
  year={2025}
}

@article{hong2025glm,
  title={Glm-4.5 v and glm-4.1 v-thinking: Towards versatile multimodal reasoning with scalable reinforcement learning},
  author={Hong, Wenyi and Yu, Wenmeng and Gu, Xiaotao and Wang, Guo and Gan, Guobing and Tang, Haomiao and Cheng, Jiale and Qi, Ji and Ji, Junhui and Pan, Lihang and others},
  journal={arXiv preprint arXiv:2507.01006},
  year={2025}
}

@article{cheng2025glyph,
  title={Glyph: Scaling context windows via visual-text compression},
  author={Cheng, Jiale and Liu, Yusen and Zhang, Xinyu and Fei, Yulin and Hong, Wenyi and Lyu, Ruiliang and Wang, Weihan and Su, Zhe and Gu, Xiaotao and Liu, Xiao and others},
  journal={arXiv preprint arXiv:2510.17800},
  year={2025}
}

@article{wei2025deepseek,
  title={Deepseek-ocr: Contexts optical compression},
  author={Wei, Haoran and Sun, Yaofeng and Li, Yukun},
  journal={arXiv preprint arXiv:2510.18234},
  year={2025}
}

@article{lu2024deepseek,
  title={Deepseek-vl: towards real-world vision-language understanding},
  author={Lu, Haoyu and Liu, Wen and Zhang, Bo and Wang, Bingxuan and Dong, Kai and Liu, Bo and Sun, Jingxiang and Ren, Tongzheng and Li, Zhuoshu and Yang, Hao and others},
  journal={arXiv preprint arXiv:2403.05525},
  year={2024}
}

@article{li2023seed,
  title={Seed-bench: Benchmarking multimodal llms with generative comprehension},
  author={Li, Bohao and Wang, Rui and Wang, Guangzhi and Ge, Yuying and Ge, Yixiao and Shan, Ying},
  journal={arXiv preprint arXiv:2307.16125},
  year={2023}
}

@inproceedings{liu2024mmbench,
  title={Mmbench: Is your multi-modal model an all-around player?},
  author={Liu, Yuan and Duan, Haodong and Zhang, Yuanhan and Li, Bo and Zhang, Songyang and Zhao, Wangbo and Yuan, Yike and Wang, Jiaqi and He, Conghui and Liu, Ziwei and others},
  booktitle={European conference on computer vision},
  pages={216--233},
  year={2024},
  organization={Springer}
}

@article{xing2025vision,
  title={Vision-centric token compression in large language model},
  author={Xing, Ling and Wang, Alex Jinpeng and Yan, Rui and Shu, Xiangbo and Tang, Jinhui},
  journal={arXiv preprint arXiv:2502.00791},
  year={2025}
}

@article{wang2024leveraging,
  title={Leveraging visual tokens for extended text contexts in multi-modal learning},
  author={Wang, Alex Jinpeng and Li, Linjie and Lin, Yiqi and Li, Min and Wang, Lijuan and Shou, Mike Zheng},
  journal={Advances in Neural Information Processing Systems},
  volume={37},
  pages={14325--14348},
  year={2024}
}

@inproceedings{kesen2025multilingual,
  title={Multilingual Pretraining for Pixel Language Models},
  author={Kesen, Ilker and Lotz, Jonas F and Ziegler, Ingo and Rust, Phillip and Elliott, Desmond},
  booktitle={Proceedings of the 2025 Conference on Empirical Methods in Natural Language Processing},
  pages={29582--29599},
  year={2025}
}

@inproceedings{liu2024improved,
  title={Improved baselines with visual instruction tuning},
  author={Liu, Haotian and Li, Chunyuan and Li, Yuheng and Lee, Yong Jae},
  booktitle={Proceedings of the IEEE/CVF conference on computer vision and pattern recognition},
  pages={26296--26306},
  year={2024}
}

@article{liu2023visual,
  title={Visual instruction tuning},
  author={Liu, Haotian and Li, Chunyuan and Wu, Qingyang and Lee, Yong Jae},
  journal={Advances in neural information processing systems},
  volume={36},
  pages={34892--34916},
  year={2023}
}

@article{tong2024cambrian,
  title={Cambrian-1: A fully open, vision-centric exploration of multimodal llms},
  author={Tong, Shengbang and Brown, Ellis and Wu, Penghao and Woo, Sanghyun and Middepogu, Manoj and Akula, Sai C and Yang, Jihan and Yang, Shusheng and Iyer, Adithya and Pan, Xichen and others},
  journal={Advances in Neural Information Processing Systems},
  volume={37},
  pages={87310--87356},
  year={2024}
}

@inproceedings{chen2024internvl,
  title={Internvl: Scaling up vision foundation models and aligning for generic visual-linguistic tasks},
  author={Chen, Zhe and Wu, Jiannan and Wang, Wenhai and Su, Weijie and Chen, Guo and Xing, Sen and Zhong, Muyan and Zhang, Qinglong and Zhu, Xizhou and Lu, Lewei and others},
  booktitle={Proceedings of the IEEE/CVF conference on computer vision and pattern recognition},
  pages={24185--24198},
  year={2024}
}

@inproceedings{tong2024eyes,
  title={Eyes wide shut? exploring the visual shortcomings of multimodal llms},
  author={Tong, Shengbang and Liu, Zhuang and Zhai, Yuexiang and Ma, Yi and LeCun, Yann and Xie, Saining},
  booktitle={Proceedings of the IEEE/CVF conference on computer vision and pattern recognition},
  pages={9568--9578},
  year={2024}
}

@article{jiang2024mantis,
  title={Mantis: Interleaved multi-image instruction tuning},
  author={Jiang, Dongfu and He, Xuan and Zeng, Huaye and Wei, Cong and Ku, Max and Liu, Qian and Chen, Wenhu},
  journal={arXiv preprint arXiv:2405.01483},
  year={2024}
}

@inproceedings{dong2025insight,
  title={Insight-v: Exploring long-chain visual reasoning with multimodal large language models},
  author={Dong, Yuhao and Liu, Zuyan and Sun, Hai-Long and Yang, Jingkang and Hu, Winston and Rao, Yongming and Liu, Ziwei},
  booktitle={Proceedings of the Computer Vision and Pattern Recognition Conference},
  pages={9062--9072},
  year={2025}
}

@inproceedings{lee2023pix2struct,
  title={Pix2struct: Screenshot parsing as pretraining for visual language understanding},
  author={Lee, Kenton and Joshi, Mandar and Turc, Iulia Raluca and Hu, Hexiang and Liu, Fangyu and Eisenschlos, Julian Martin and Khandelwal, Urvashi and Shaw, Peter and Chang, Ming-Wei and Toutanova, Kristina},
  booktitle={International Conference on Machine Learning},
  pages={18893--18912},
  year={2023},
  organization={PMLR}
}

@article{li2025text,
  title={Text or pixels? it takes half: On the token efficiency of visual text inputs in multimodal llms},
  author={Li, Yanhong and Lan, Zixuan and Zhou, Jiawei},
  journal={arXiv preprint arXiv:2510.18279},
  year={2025}
}

@article{yang2024vision,
  title={Vision model pre-training on interleaved image-text data via latent compression learning},
  author={Yang, Chenyu and Zhu, Xizhou and Zhu, Jinguo and Su, Weijie and Wang, Junjie and Dong, Xuan and Wang, Wenhai and Lu, Lewei and Li, Bin and Zhou, Jie and others},
  journal={Advances in Neural Information Processing Systems},
  volume={37},
  pages={23912--23938},
  year={2024}
}

@inproceedings{leng2024mitigating,
  title={Mitigating object hallucinations in large vision-language models through visual contrastive decoding},
  author={Leng, Sicong and Zhang, Hang and Chen, Guanzheng and Li, Xin and Lu, Shijian and Miao, Chunyan and Bing, Lidong},
  booktitle={Proceedings of the IEEE/CVF Conference on Computer Vision and Pattern Recognition},
  pages={13872--13882},
  year={2024}
}

@article{schrodi2024two,
  title={Two effects, one trigger: On the modality gap, object bias, and information imbalance in contrastive vision-language models},
  author={Schrodi, Simon and Hoffmann, David T and Argus, Max and Fischer, Volker and Brox, Thomas},
  journal={arXiv preprint arXiv:2404.07983},
  year={2024}
}

@article{liang2022mind,
  title={Mind the gap: Understanding the modality gap in multi-modal contrastive representation learning},
  author={Liang, Victor Weixin and Zhang, Yuhui and Kwon, Yongchan and Yeung, Serena and Zou, James Y},
  journal={Advances in Neural Information Processing Systems},
  volume={35},
  pages={17612--17625},
  year={2022}
}

@inproceedings{sun2024aligning,
  title={Aligning large multimodal models with factually augmented rlhf},
  author={Sun, Zhiqing and Shen, Sheng and Cao, Shengcao and Liu, Haotian and Li, Chunyuan and Shen, Yikang and Gan, Chuang and Gui, Liangyan and Wang, Yu-Xiong and Yang, Yiming and others},
  booktitle={Findings of the Association for Computational Linguistics: ACL 2024},
  pages={13088--13110},
  year={2024}
}

@article{wang2024mmlu,
  title={Mmlu-pro: A more robust and challenging multi-task language understanding benchmark},
  author={Wang, Yubo and Ma, Xueguang and Zhang, Ge and Ni, Yuansheng and Chandra, Abhranil and Guo, Shiguang and Ren, Weiming and Arulraj, Aaran and He, Xuan and Jiang, Ziyan and others},
  journal={arXiv preprint arXiv:2406.01574},
  year={2024}
}

@article{cobbe2021gsm8k,
  title={Training Verifiers to Solve Math Word Problems},
  author={Cobbe, Karl and Kosaraju, Vineet and Bavarian, Mohammad and Chen, Mark and Jun, Heewoo and Kaiser, Lukasz and Plappert, Matthias and Tworek, Jerry and Hilton, Jacob and Nakano, Reiichiro and Hesse, Christopher and Schulman, John},
  journal={arXiv preprint arXiv:2110.14168},
  year={2021}
}

@article{chen2021codex,
  title={Evaluating Large Language Models Trained on Code},
  author={Mark Chen and Jerry Tworek and Heewoo Jun and Qiming Yuan and Henrique Ponde de Oliveira Pinto and Jared Kaplan and Harri Edwards and Yuri Burda and Nicholas Joseph and Greg Brockman and Alex Ray and Raul Puri and Gretchen Krueger and Michael Petrov and Heidy Khlaaf and Girish Sastry and Pamela Mishkin and Brooke Chan and Scott Gray and Nick Ryder and Mikhail Pavlov and Alethea Power and Lukasz Kaiser and Mohammad Bavarian and Clemens Winter and Philippe Tillet and Felipe Petroski Such and Dave Cummings and Matthias Plappert and Fotios Chantzis and Elizabeth Barnes and Ariel Herbert-Voss and William Hebgen Guss and Alex Nichol and Alex Paino and Nikolas Tezak and Jie Tang and Igor Babuschkin and Suchir Balaji and Shantanu Jain and William Saunders and Christopher Hesse and Andrew N. Carr and Jan Leike and Josh Achiam and Vedant Misra and Evan Morikawa and Alec Radford and Matthew Knight and Miles Brundage and Mira Murati and Katie Mayer and Peter Welinder and Bob McGrew and Dario Amodei and Sam McCandlish and Ilya Sutskever and Wojciech Zaremba},
  year={2021},
  eprint={2107.03374},
  archivePrefix={arXiv},
  primaryClass={cs.LG}
}

@inproceedings{jain2025livecodebench,
  title={Livecodebench: Holistic and contamination free evaluation of large language models for code},
  author={Jain, Naman and Gu, Alex and Li, Wen-Ding and Yan, Fanjia and Zhang, Tianjun and Wang, Sida and Solar-Lezama, Armando and Sen, Koushik and Stoica, Ion},
  booktitle={International Conference on Learning Representations},
  volume={2025},
  pages={58791--58831},
  year={2025}
}

@misc{zhou2023ifeval,
      title={Instruction-Following Evaluation for Large Language Models}, 
      author={Jeffrey Zhou and Tianjian Lu and Swaroop Mishra and Siddhartha Brahma and Sujoy Basu and Yi Luan and Denny Zhou and Le Hou},
      year={2023},
      eprint={2311.07911},
      archivePrefix={arXiv},
      primaryClass={cs.CL},
      url={https://arxiv.org/abs/2311.07911}, 
}

%%%%%%%%%%%%%%%%%%%%%%%%%%%%%%%%%%%%%%%%%%%%%%%%%%%%%%%%%%%%
\newpage
% =====================================================================
%  appendix.tex
%  ---------------------------------------------------------------------
%  Technical Appendix for the LoMo paper (NeurIPS 2026 submission).
%
%  Usage:
%    In your main .tex file, after \bibliography{...} and before the
%    NeurIPS Paper Checklist, simply add:
%
%        \input{appendix.tex}
%
%    This file begins with \appendix, so all subsequent \section{}s
%    are auto-lettered (A, B, C, ...) and \subsection{}s become
%    A.1, A.2, etc. Do NOT call \appendix again in the main file.
%
%  Notes:
%    - Keep the NeurIPS Paper Checklist AFTER this file is included.
%    - The Checklist must remain unnumbered (\section*{}), not part of
%      the \appendix-lettered sequence.
% =====================================================================

\appendix

\section{Pure-Text Capability Analysis}
\label{app:pure-text}

To verify the effect of LoMo on pure-text capabilities, we evaluate Standard SFT and LoMo on five pure-text benchmarks: MMLU-Pro~\cite{wang2024mmlu}, GSM8K~\cite{cobbe2021gsm8k}, HumanEval~\cite{chen2021codex}, LiveCodeBench V6~\cite{jain2025livecodebench}, and IFEval~\cite{zhou2023ifeval}. As shown in Table~\ref{tab:pure-text}, LoMo matches or slightly exceeds Standard SFT on every benchmark, with average gains of $+0.28$ and $+0.58$ on the two backbones. On Qwen3.5-9B, the pure-text IFEval gain ($+2.59$) is in the same direction as the multimodal MM-IFEval gain ($+5.49$, Table~\ref{tab:main}). These results show that LoMo improves multimodal performance without compromising pure-text capabilities, achieving small average gains on both backbones.
\begin{table*}[h]
\centering
\small
\setlength{\tabcolsep}{3.5pt}
\renewcommand{\arraystretch}{1.15}
\caption{Pure-text capability sanity check on
LLaVA-OV1.5-8B and Qwen3.5-9B. We evaluate both Standard SFT
and LoMo on five widely used pure-text benchmarks covering
general knowledge, mathematical reasoning, code generation,
and instruction following. $\Delta$ denotes the absolute change
of LoMo over Standard SFT; \textcolor{darkgreen}{$\uparrow$} and
\textcolor{red}{$\downarrow$} indicate gains and drops. Benchmark abbreviations: MMLU-P: MMLU-Pro,
LCB-V6: LiveCodeBench V6.}
\vspace{2mm}
\label{tab:pure-text}
\resizebox{0.8\textwidth}{!}{%
\begin{tabular}{l l c cc cc | c}
\toprule
\multirow{2}{*}{Model} & \multirow{2}{*}{Setting}
 & Reasoning & \multicolumn{2}{c}{Coding} & Math & Instruct.
 & \multirow{2}{*}{Avg.} \\
\cmidrule(lr){3-3}\cmidrule(lr){4-5}\cmidrule(lr){6-6}\cmidrule(lr){7-7}
 & & MMLU-P & HumanEval & LCB-V6 & GSM8K & IFEval & \\
\midrule
\multirow{3}{*}{LLaVA-OV1.5-8B}
 & Standard SFT
 & 62.58 & 82.62 & 29.19 & 92.87 & 75.42
 & 68.54 \\
 & \textbf{+ LoMo}
 & \textbf{62.27} & \textbf{82.93} & \textbf{29.91}
 & \textbf{93.40} & \textbf{75.60}
 & \textbf{68.82} \\
 & $\Delta$
 & \dn{0.31} & \up{0.31} & \up{0.72}
 & \up{0.53} & \up{0.18}
 & \up{0.28} \\
\cmidrule(lr){1-8}
\multirow{3}{*}{Qwen3.5-9B}
 & Standard SFT
 & 70.74 & 71.34 & 33.29 & 95.38 & 72.27
 & 68.60 \\
 & \textbf{+ LoMo}
 & \textbf{71.16} & \textbf{71.95} & \textbf{33.43}
 & \textbf{94.54} & \textbf{74.86}
 & \textbf{69.19} \\
 & $\Delta$
 & \up{0.42} & \up{0.61} & \up{0.14}
 & \dn{0.84} & \up{2.59}
 & \up{0.58} \\
\bottomrule
\end{tabular}}
% \vspace{4mm}
\end{table*}

\begin{algorithm}[H]
\footnotesize
\SetCommentSty{textnormal}
\DontPrintSemicolon
\caption{The pipeline of LoMo.}           % 短 caption,一行搞定
\label{alg:LoMo}
\KwIn{Text-only instance $(x,a)$}
\KwOut{Interleaved instance $(U,a)$}
\tcp{Stage 1: Span localization}
\eIf{$\mathrm{SentenceCount}(x) \leq 3$}{
  $(x_{\mathrm{pre}}, x_{\mathrm{mid}}, x_{\mathrm{suf}}) \gets (\emptyset, x, \emptyset)$\;
}{
  $\mathcal{C} \gets \mathrm{FormulaAwareChunk}(x)$\;
  $(x_{\mathrm{pre}}, x_{\mathrm{mid}}, x_{\mathrm{suf}}) \gets \mathrm{ExtractMiddle}(\mathcal{C})$\;
}
\tcp{Stage 2: Visual rendering}
\eIf{$\mathrm{ContainsMath}(x_{\mathrm{mid}})$}{
  $I \gets \mathrm{LaTeXRender}(x_{\mathrm{mid}})$\;
  \lIf{$I = \mathrm{None}$}{$I \gets \mathrm{TextRender}(x_{\mathrm{mid}})$}
}{
  $I \gets \mathrm{TextRender}(x_{\mathrm{mid}})$\;
}
$I \gets \mathrm{TrimMargin}(I)$\;
\tcp{Stage 3: Perceptual distortion}
$\mathcal{O} \gets \{\mathrm{Clean}, \mathrm{Rotate}, \mathrm{Blur}, \mathrm{ShadowOrStain}, \mathrm{Wave}\}$\;
$I' \gets \mathrm{SubOps}(o)(I),\ \text{where } o \sim \mathrm{Categorical}(\mathcal{O})$\;
\Return $((x_{\mathrm{pre}}, I', x_{\mathrm{suf}}), a)$\;
\end{algorithm}

% =====================================================================
\section{Implementation Details}
\label{app:impl}
% =====================================================================

\subsection{Compute Resources}
\label{sec:compute}

Image rendering for the LoMo data construction pipeline was performed 
on two CPU servers with 128 cores each, taking approximately 20 hours 
in total. All model training and evaluation experiments were conducted 
on a single node equipped with 8 NVIDIA H200 GPUs.

\subsection{Training Data Construction}
\label{app:impl-data}

Our training corpus combines the original LLaVA-OneVision 1.5 instruct 
data with our LoMo-augmented rendered-text data. Specifically, we 
sample 2M multimodal and 2M text-only instances, with 50\% of the 
latter rendered via LoMo. For a fair comparison, the standard SFT 
baseline is trained on the same 4M instances without any LoMo 
rendering. The rendered-text data is constructed by rendering 
text question prompts from the pure-text corpus into images. Rendering 
is performed with the Python-based text renderer with LaTeX support 
enabled. The plain text uses font size $20$ with line height $22$, 
while mathematical expressions are typeset in common Latin Modern Math 
at size $26$. Rendered images are retained at their native resolution 
based on the length of text. The resulting rendered images are 
inserted back into multimodal SFT examples in place of the 
corresponding text spans.

\subsection{Training Setup}
\label{app:impl-train}

We fine-tune LLaVA-OneVision-1.5-8B-Base and Qwen3.5-9B-Base using LLaMA-Factory under a
standard supervised fine-tuning regime.  We adopt a
maximum sequence length of $32{,}768$ tokens, and a maximum image resolution of $2{,}560{,}000$ pixels. Training employs FlashAttention 2, bf16 precision and DeepSpeed ZeRO Stage~1, with liger kernels and sequence packing for throughput. The learning rate follows a cosine schedule from $4\mathrm{e}{-}5$ to $1\mathrm{e}{-}6$ with a warmup ratio of $0.002$. We use a per-device batch size of $1$ with $4$ gradient accumulation steps. The language model and the multimodal projector are updated, while the vision tower remains frozen throughout training. We optimize with AdamW ($\beta_1{=}0.9$, $\beta_2{=}0.99$, weight decay $0.01$) for one epoch.

\subsection{Standard Evaluation Protocol}
\label{app:impl-eval-std}

We evaluate all models with EvalScope under its standard protocol. 
The process of model prediction is deterministic with temperature $0$ and a maximum context 
length of $32$K tokens.

\subsection{Rendered Evaluation Protocol}
\label{app:impl-eval-rend}

Under the rendered evaluation protocol, each textual question is rendered into an 
image using a python-based renderer and font configuration with font size $20$ and line height $22$, and is then used in place of the original text question.  All other settings remain identical to the standard evaluation protocol.

\section{Limitations}
While LoMo delivers consistent improvements, several aspects remain beyond the scope of this work. We apply LoMo only during the SFT stage, leaving its integration with pre-training or RL-based post-training to future exploration. Our span localization adopts a block-level middle-span heuristic; exploring more fine-grained strategies such as difficulty-aware or curriculum-style selection may yield further gains. Finally, due to compute constraints, we validate LoMo on two backbones at the 8B–9B scale, and verifying its behavior on substantially larger models is left to future work.

\section{Broader Impacts}
\label{sec:impact}

LoMo strengthens cross-modal alignment in Vision-Language Models without modifying model architectures 
or introducing inference overhead. By treating text and images as 
semantically interchangeable carriers, it improves model robustness 
in real-world scenarios where information appears in mixed 
visual-textual forms, benefiting applications such as document 
understanding, assistive reading tools, and educational platforms 
that integrate handwritten or printed materials. As LoMo builds on 
pre-trained VLMs, it inherits the biases and failure modes of these 
foundation models, and we encourage practitioners to follow 
established VLM safety practices when deploying LoMo-trained models 
in high-stakes domains.

% =====================================================================
% Optional follow-up appendix sections. Uncomment and fill in as needed.
% =====================================================================

% \section{Additional Experimental Results}
% \label{app:more-results}
%
% \subsection{Span Position Ablation}
% \label{app:span-position}
%
% \subsection{Rewrite Ratio Sweep}
% \label{app:rewrite-ratio}
%
% \subsection{Pure-Text Capability Sanity Check}
% \label{app:pure-text}

% \section{Qualitative Examples}
% \label{app:examples}
%
% \subsection{LoMo-augmented Training Samples}
% \label{app:examples-train}
%
% \subsection{Rendered Question Examples per Benchmark}
% \label{app:examples-eval}

% End of appendix.tex
 
%%%%%%%%%%%%%%%%%%%%%%%%%%%%%%%%%%%%%%%%%%%%%%%%%%%%%%%%%%%%

% \newpage
% \input{checklist.tex}

\end{document}